\documentclass{article}
\usepackage{ijcai20}

\usepackage{times}

\usepackage{soul}
\usepackage{url}
\usepackage[utf8]{inputenc}
\usepackage[small]{caption}
\usepackage{graphicx}
\usepackage[export]{adjustbox}
\usepackage{amsmath}
\usepackage{booktabs}
\usepackage{algorithm}
\usepackage{algorithmic}
\usepackage{float}
\usepackage{amssymb,amsthm,mathrsfs}
\usepackage{bm}
\usepackage{pifont}
\usepackage{multirow}
\usepackage{subfigure}
\usepackage{arydshln} 
\usepackage{verbatim}

\usepackage[dvipsnames,svgnames,x11names]{xcolor}

\usepackage{color}
\definecolor{citecolor}{RGB}{119,185,0} 
\usepackage[pagebackref=false,breaklinks=true,letterpaper=true,colorlinks,citecolor=citecolor,bookmarks=false]{hyperref}

\def\eg{\emph{e.g.}} 
\def\ie{\emph{i.e.}} 
\def\etal{\emph{et~al.}} 

\newlength\savewidth\newcommand\shline{\noalign{\global\savewidth\arrayrulewidth
  \global\arrayrulewidth 1pt}\hline\noalign{\global\arrayrulewidth\savewidth}}

\title{Progressive Local Filter Pruning for Image Retrieval Acceleration}
\author{
Xiaodong Wang$^{1,2}$\footnote{Work done during the visiting at University of Technology Sydney}\and
Zhedong Zheng$^2$\and
Yang He$^{2}$\and
Fei Yan$^1$ \and 
Zhiqiang Zeng$^1$ \and
Yi Yang$^{2}$ 
\affiliations
$^1$Xiamen University of Technology \\
$^2$ReLER, University of Technology Sydney\\
\emails
xdwangjsj@xmut.edu.cn, \
\{zhedong.zheng,  yang.he-1\}@student.uts.edu.au, \\
\{fyan, zqzeng\}@xmut.edu.cn, yi.yang@uts.edu.au
}

\begin{document}

\maketitle

\begin{abstract}
This paper focuses on network pruning for image retrieval acceleration. Prevailing image retrieval works target at the discriminative feature learning, while little attention is paid to how to accelerate the model inference, which should be taken into consideration in real-world practice.
The challenge of pruning image retrieval models is that the middle-level feature should be preserved as much as possible. Such different requirements of the retrieval and classification model make the traditional pruning methods not that suitable for our task.
To solve the problem, we propose a new Progressive Local Filter Pruning (PLFP) method for image retrieval acceleration. Specifically, layer by layer, we analyze the \textit{local} geometric properties of  each filter and select the one that can be replaced by the neighbors. Then we \textit{progressively} prune the filter by gradually changing the filter weights. In this way, the representation ability of the model is preserved.
To verify this, we evaluate our method on two widely-used image retrieval datasets, \ie, Oxford5k and Paris6K, and one person re-identification dataset, \ie, Market-1501. 
The proposed method arrives with superior performance to the conventional pruning methods, suggesting the effectiveness of the proposed method for image retrieval. 
\end{abstract}

\section{Introduction}

\begin{figure}[t]
\center{
\subfigure[Criteria for filter pruning: \textbf{(left)} global center-based  methods, (\textbf{right}) our method. ]{
\begin{minipage}[c]{1\linewidth}
\center{\includegraphics[width=0.8\textwidth]{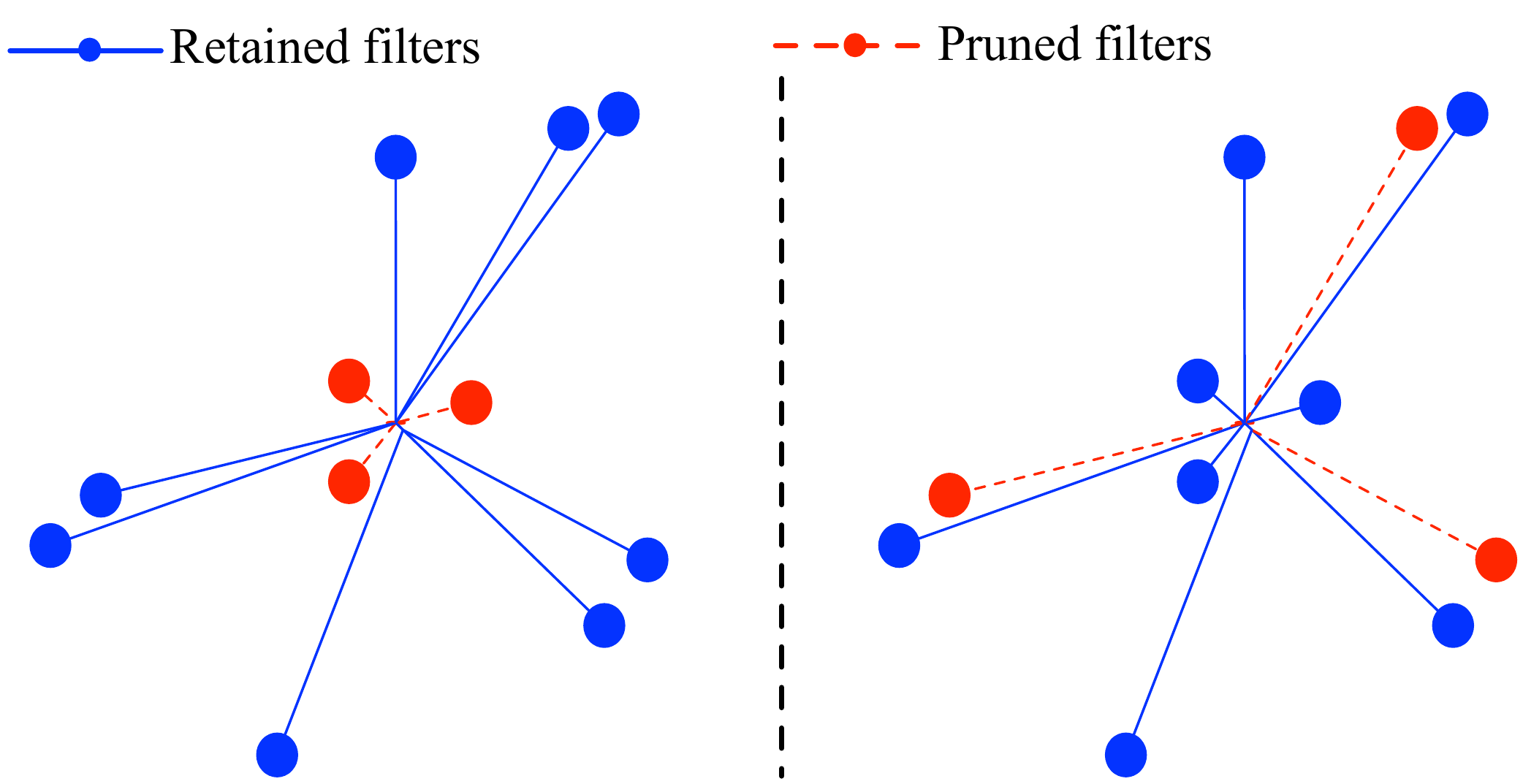}}
\end{minipage} 
}
\center{
\subfigure[Filter distributions of different methods: ~(\textbf{left}) original filter distribution, (\textbf{middle}) global center-based  methods, (\textbf{right}) our method. ]{
\begin{minipage}[c]{1\linewidth}
\center{\includegraphics[width=\textwidth]{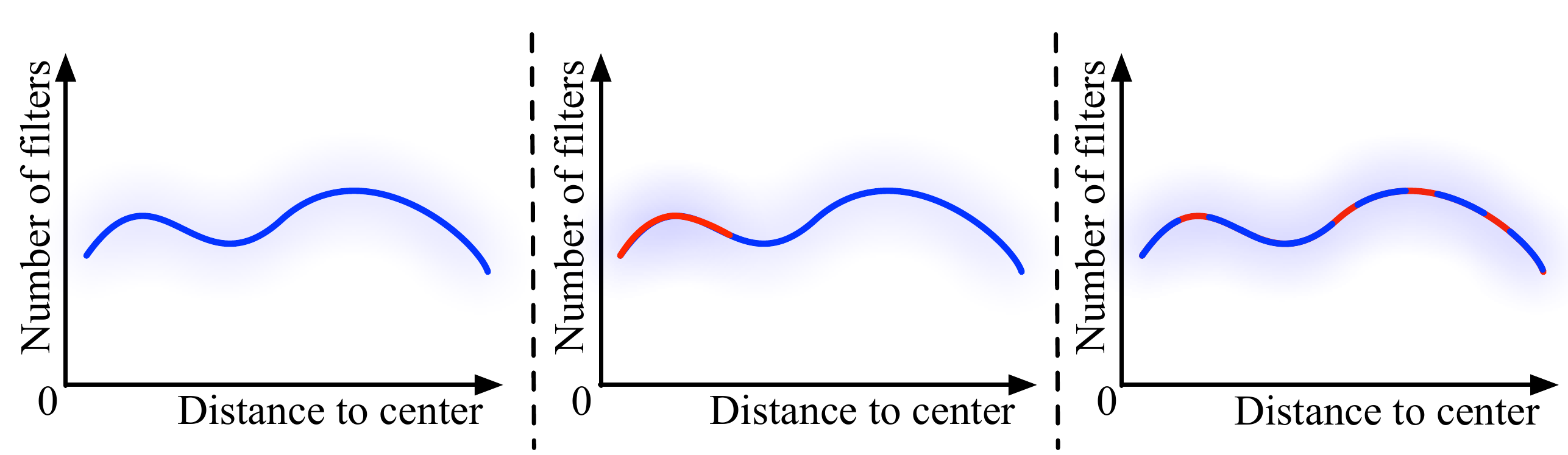}}
\end{minipage} 
}
}
\vspace{-.1in}
\caption{A comparison between the conventional global  center-based methods and the proposed method. In (a) the circles denote the filters, where lines denote the distance of filters to the geometric center.  For global center-based  methods, only the filters (with long lines) far away  from the geometric center are kept, while our method prunes filters according to the local relationships to the neighbors. (b) shows the three distributions of different methods.  The one generated by global center-based  methods (middle) is quite different from the original filter distribution (left), while our method (right) properly maintain the  original filter distribution.}
\label{fig1}
}
\end{figure}

Image representation learned by Convolutional Neural Network (CNN) has become dominant in image retrieval, due to the discriminative power. However, training and testing the CNN-based model usually demands expensive computation resources for the fast calculation. It remains challenging for CNN-based applications to the platforms with limited resources, \eg, cellphones and self-driving cars.  
To accelerate the inference, researchers resort to simplifying the CNN-based model to find a trade-off between performance and efficiency~\cite{Li2016PruningFF,He2017ChannelPF,he2018SFP,Jonathan2018Lottery}. However, the existing network pruning methods mostly target the image classification problem. Pruning image retrieval models remains underexplored.

In this paper, we intend to fill the gap, and focus on network pruning methods for image retrieval acceleration. 
Compared with the image classification model, of which the output is a discrete category, the image retrieval task aims at learning continuous features. Therefore, image retrieval model is sensitive to the pruning operation in nature, which would lead to unexpected feature distribution changes. This paper addresses two main challenges in pruning image retrieval models.
First, \textbf{it is critical to preserve the geometry structure of filters in the original model}, \ie, the filter relation of the pruned model should be the same as the original model. 
The existing methods mainly adopt the global center-based pruning approaches, which rank the filters according to the distance to the geometric center \cite{Li2016PruningFF,he2018SFP,he2019FPGM}. However, as shown in~Fig.\ref{fig1}(b), they suffer from destroying the original filter distribution.  The pruning procedure of our method and global center-based methods are illustrated in Fig.~\ref{fig1}(a). We could observe that the global center-based methods tend to prune filters closed to the geometric center, resulting in a great change in filter distributions (Fig.\ref{fig1}(b), middle), while our method (Fig.\ref{fig1}(b), right) can properly maintain the original filter distributions (Fig.\ref{fig1}(b), left). 

Second, \textbf{maintaining the representation ability during pruning remains challenging}.  The ``lottery'' mechanism \cite{Jonathan2018Lottery,sun2019learning} proposes to drop specific channels and argues that the key is in the weight initialization. 
Iterative training and rewinding is necessary for this line of methods. In this way, the models of different iterations usually have totally different feature spaces from the original model.
In parallel, some pruning techniques \cite{he2018SFP,he2019FPGM} argue that the devil is in the training process and drop the weight by deploying dynamic masks. The masks directly set the redundant filters to zeros, which also affects the learned feature space largely.
In this paper, we start from a well-trained model and remove redundant filters by decreasing the weight scale gradually. It could be viewed as one mild pruning method instead of setting the filter to zeros immediately. The model representation ability, therefore, is preserved, especially for the high-proportion pruning demands.

In an attempt to overcome the above-mentioned challenges, we propose Progressive Local Filter Pruning (PLFP) to accelerate CNN models for image retrieval. 
1) To maintain the distribution of the original model, we propose the local filter pruning criterion, which is different from conventional global-based criteria. The proposed method preserves the filter distribution of the original model. 
2) Different from existing methods, we do not prune the model with zero masks or retrain the model from lottery initialization. Instead, we leverage the well-trained model and adopt one progressive pruning policy by changing the weight scales.
To summarize, this paper has the following contributions.
\begin{itemize}
\item 
We propose to leverage the local geometric properties of the filter distribution. Thanks to the local similarity, we iteratively find the redundant filters which could be replaced by the neighbor filters. Therefore, the proposed pruning method could preserve the geometry structure of filters in the original model, even when a large proportion of filters is removed.
\item As a minor contribution, we also propose a simple filter weight decreasing strategy to prune the filters progressively. Albeit simple, the progressive pruning allows the pruned model to recover the representation capability, especially when dropping a large proportion of filters. 
The ablation study also verifies the effectiveness of the progressing pruning.
\item The extensive results on several benchmark datasets demonstrate the effectiveness of the  proposed method. After reducing more than 57\% FLOPs, the pruned model still achieves 76.20\%, 73.18\%  $mAP$ on Oxford5k and Paris6k, respectively. Furthermore, we reduce 88.9\% FLOPs with only 4.2\% $top\text{-}5$  error increase on  Market-1501.
 \end{itemize}
 
\begin{figure*}[t]
\begin{minipage}[c]{1\linewidth}
\center{\includegraphics[width=1\textwidth]{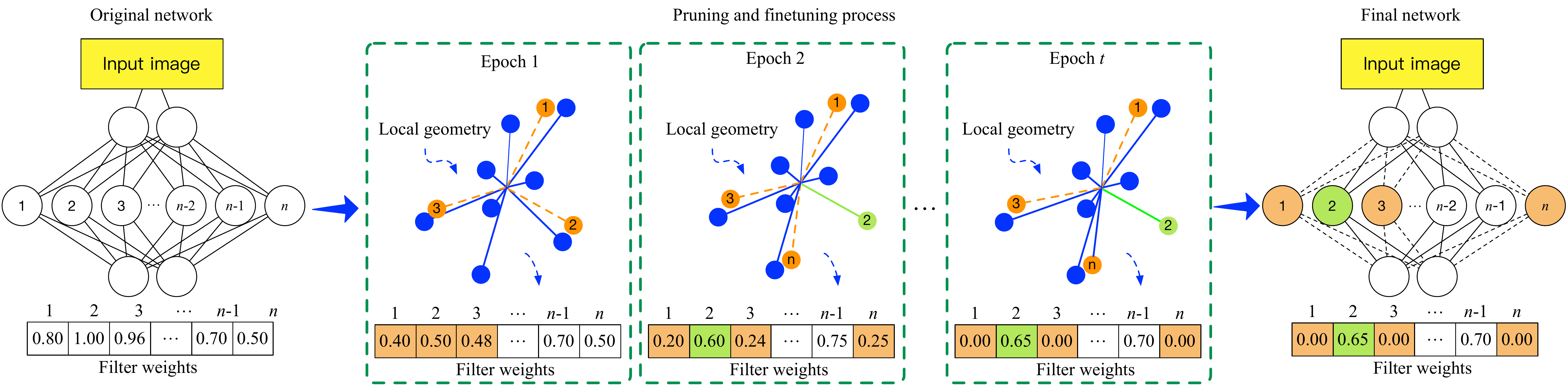}}
\end{minipage} 
\vspace{-.1in}
\caption[]{The geometry interpretation of the proposed method. Given a pretrained model, for a specific layer (the second layer in this example), we first calculate the local geometric properties for every filter and select redundant filters (orange filled circles) with small local power (more details are in Section \protect\ref{sec:filter_decrease}). Then we deploy the filter weight decreasing policy to adjust the scale of the selected filters (orange filled squares). %(by multiplying $\gamma=0.5$ in this example). 
We iteratively re-sample candidate filters and fine-tune the model to give the wrong-selected filters chances to be rectified (green filled square), till the weight of redundant filters gradually converges to zeros. Finally, we could obtain a slim model after removing the zero-weight filters.} %, which preserves the original representation ability.}
\label{fig_arch}
\end{figure*}

\section{Related Work}

\textbf{Image Retrieval.} Recently, Convolutional Neural Network (CNN) has been widely applied to extract the visual features \cite{Babenko2014NeuralCF,toliasICLR2016,Babenko2015AggregatingLD}.
\cite{Babenko2014NeuralCF} proposes to leverage the well-trained image classification model and fine-tune the model to preserve the common knowledge for image retrieval tasks.
The CNN also demands large-scale training data. Therefore, \cite{Radenovi2018Retrieval} proposes to utilize the 3D model to help data collection, which could greatly save the cost of manual annotation. 
Besides, on some sub-tasks of image retrieval, such as person re-identification \cite{ZhengAxiv16,ZhengTOMM2017}, CNN also largely improves the preformance. 
Although the CNN-based feature has shown the effectiveness on the image retrieval tasks, little attention is paid to the computational cost of the image retrieval model. 
%The feature extraction time  highly depends on the model size.

\textbf{Network Acceleration.} There are several previous works on accelerating CNNs, such as  matrix decomposition \cite{Jaderberg2014}, quantization methods \cite{Gong2014CompressingDC,Han2015DeepCC,rastegariECCV16}, knowledge distillation \cite{DistrillNIPS2015,Chen2016KnowledgeTransfer,DistrillNIPS2018}, and network pruning methods~\cite{he2018SFP}.
Among them, pruning methods have attracted much attention as they can properly reduce computation cost and accelerate the inference by removing unnecessary filters or connections.
According to %whether deleting connections permanent during pruning
the pruning policy, previous pruning methods can be roughly categorized into hard-pruning and soft-pruning methods.  
The early research on hard-pruning starts at the brain damage \cite{LeCunNIPS1989} and brain surgeon \cite{NIPHassibiS1993} techniques.  \cite{HanPTD2015} removes connections with low-weight, while \cite {Li2016PruningFF} determines the importance of filters by the $l_1$-norm values. 
Another line of pruning method, soft-pruning, is a sort of training compatibility method. Soft-pruning does not drop the pruned filters immediately but utilizes a dynamic mask to set candidate filters to zeros temporally. The critical filters could be recovered later \cite{he2018SFP}. The method also could be smoothly embedded into the training framework of traditional CNN models. 
However, the soft-pruning methods still drop the filter via setting the weights to zeros and may largely compromise the well-trained weight from the original model, especially when pruning a large number of filters.

\section{Methodology}

\subsection{Preliminaries}
As shown in Fig.~{\ref{fig_arch}}, we briefly illustrate the pruning process for a single convolutional layer. Our method first selects the redundant filters by the local geometry relationships. More details are provided in Section \ref{sec:local}. The candidate filters are not immediately dropped. Instead, we adopt the weight decreasing policy in Section \ref{sec:filter_decrease} to decrease the weight scale of selected filters. In this way, we progressively decrease the impact of the candidate filters on the network. Meanwhile, some wrong-selected filters could have chances of recovering to the original scale in the following training process. 

\subsection{Local Filter Selection } \label{sec:local}

\textbf{Formulation.} We denote $W^l \in \mathbb{R}^{C_{out}^l\times  C_{in}^l \times K^l \times K^l }$ as the weight matrix of the $l$-th convolutional layer, where $C^l_{out}$ and $C^l_{in}$ denote the number of the output channels and the input channels, respectively. $K^l$ is the kernel size. We intend to find redundant filters, which could be replaced by other filters. To this end, we reshape the weights $W^l$ as $C_{out}^l\times  L^l$ with $L^l = C_{in}^l \times K^l \times K^l $. We could view the weights of the $l$-th layer $W^l$ as a set of filters $W^l = [w^l_1,w^l_2,\cdots , w^l_{C^l_{out}} ]$, where $w^l_{i}$ is the $i$-th filter with the length of $L^l$.

\noindent\textbf{Local Geometry.} We intend to find redundant filters according to the mutual information. %while maintaining the high-level semantic representation of the original model. 
Inspired by previous works on the local manifold learning~\cite{Yang2011UDFS,Li2017ijcai}, which aim at locality structure exploring, we argue that the local geometry of filters could reflect more accurate mutual information. Instead of considering the relationship between all filters, we focus on the local geometry between neighbor filters. Intuitively, if one filter shares similar local properties with the neighbors, the filter could be replaced by the neighbor filters. For the $i$-th filter $w^l_i$ of the $l$-th layer, we denote the neighbor filter set as $\mathscr{N}(w^l_i)$ consisting of the $k$ nearest neighbors of  $w^l_i$. We could formulate the local property of the filter as:
\begin{equation}
\begin{aligned} \label{eqn:local_power}
    \mathscr{D}(w^{l}_i) = \frac{1}{k}\sum\limits_{w \in \mathscr{N}(w^l_{i})} \psi(w,w^l_{i}),\\
\end{aligned}
\end{equation}
where $\psi(w,w^l_{i})$ denotes the  distance function to model the local relationships between $w$ and $w^l_{i}$. Specifically, we deploy the $\mathcal{L}_2$ distance $\psi(w,w^l_{i}) = || w - w^l_{i} ||_2 $.

Based on the previous analysis, we intend to prune filters that obtain the small local power to preserve the original filter distribution of the network. 
Concretely, given the pruning rate $P^l$ for the $l$-th layer, we want to obtain the optimal filter subset after pruning  $m=P^l \times C^l_{out}$ filters. The objective could be formulated as:
\begin{equation}
\begin{aligned} \label{eqn:obj}
        \min\limits_{s} \sum\limits_{i=1}^{C^l_{out}}  \bigg (\mathscr{D}( s_i w^{l}_i)  - \mathscr{D}(w^{l}_i) \bigg),\\
        s.t.\ s_i\in \{0,1\}, \sum\limits_{i=1}^{C^l_{out}} s_i = C^l_{out}-m, 
\end{aligned}
\end{equation}
where $s$ is a filter selection vector. If $s_i$ equals to 1, the filter $w^l_i$ will remain, otherwise, the filter is pruned. \textbf{Note that every time we change the filter selection $s$, some filters are pruned and the local neighbor relationship is also changed.} To solve the Eq.~(\ref{eqn:obj}), one naive way is to sort filters according to the local geometry distance $D(w^l_{(i)})$, and then select  $m$ filters with the smallest local power. However, this way may lead to an elimination of filters in the most dense area. Meanwhile, the filter selection $s$ may be sub-optimal. 
To optimize the objective, we propose to update the filter selection in an iterative way. Specifically, we sample one filter in $W^{l}$ according to Eq.~(\ref{eqn:obj}), re-evaluate the local power of each remaining filter, and repeat the optimization procedure till $m$ filters are sampled. If there are two or more filters obtaining the same local geometry distance. We calculate the global distance and sample the filter with the minimum global distance score. The detailed algorithm is provided in Algorithm~\ref{alg1}.

\noindent \textbf{Advantages of Local Filter Selection.} The conventional pruning methods usually apply the center-based criterion, \eg, $l_1$-norm and Euclidean distance to the clustering center. We argue that the center-based criterion may prune the critical filters which are close to the center, and change the original filter distribution. In contrast, the proposed method focuses on keeping the filter diversity as well as preserving filter distribution of the original model. We, therefore, leverage the local geometry as the indicator to search the candidate filters. The candidates with small local power could be replaced by the near neighbors. In this way, after pruning, the model could recover the original representation capability. 

\begin{algorithm}[t]
\caption{ Local Filter Selection } %($\mathcal{F}^{l}$,$P^l$,$k$ )}
\label{alg1} 

\begin{algorithmic}[1] 

\REQUIRE 

The original weight $\mathcal{W}^l\in R^{C_{out}^l\times  L^l}$, the prune rate $P^l$ for the $l$-th layer and the nearest neighbor number $k$.\\

\ENSURE  
The selected filter subset $\hat{\mathcal{T}^{l}}$  with the best local power preservation.\\

\STATE Initialize the selected subset $\hat{\mathcal{T}^{l}} \leftarrow  \lbrace \rbrace$ 
\STATE Initialize the rest subset $\dot{\mathcal{T}^{l}} \leftarrow  \lbrace w^l_1,w^l_2,\cdots,w^l_{C_{out}^l} \rbrace$
\STATE Construct an undirected graph $G = \langle V, E\rangle$ with the vertex set $V =  \lbrace w^l_1,w^l_2,\cdots,w^l_{C_{out}^l} \rbrace$ and the edge set $E$ are defined as: $\forall { v_i, v_j \in V, 1 \leq i,j \leq |V|}$, \\
$E_{ij}  \leftarrow \begin{cases} 
            \psi(v_i,v_j), &  if \ i \neq j \\ 
            0, & \text{Otherwise}. 
          \end{cases} $
\\          
\WHILE{$\arrowvert  \hat{\mathcal{T}^{l}} \arrowvert < P^l C^l_{out} $}

\STATE $i^* = \mathop{\arg\min}_{ 1 \leq i \leq \arrowvert \dot{\mathcal{T}}^l \arrowvert } \mathscr{D}(\dot{\mathcal{T}^{l}_i})$
\STATE $\mathcal{K} = \{m | \mathscr{D}(\dot{\mathcal{T}^{l}_m}) = \mathscr{D}(\dot{\mathcal{T}^{l}_{i^*}}) , 1 \leq m \leq \arrowvert \dot{\mathcal{T}}^l \arrowvert \}$

\STATE $j^{*} = \mathop{\arg\min}_{j \in \mathcal{K}} \sum_{i=1}^{\arrowvert \dot{\mathcal{T}^l} \arrowvert} E_{ji}$ \\
\STATE $\hat{\mathcal{T}^l}  \leftarrow  \hat{\mathcal{T}^l} \bigcup \mathcal{T}_{j^{*}}^l$, $\dot{\mathcal{T}^l}  \leftarrow  \dot{\mathcal{T}^l} - \mathcal{T}_{j^{*}}^l$\\ 

\STATE Delete node $v_{j^{*}}$ in $V$ and the incident edges from $G$ \\
\ENDWHILE\\

\RETURN $\hat{\mathcal{T}^l}$ 

\end{algorithmic}
\end{algorithm}

\subsection{Filter Weight Decreasing}
\label{sec:filter_decrease}
We could obtain the candidate filter subset $\hat{\mathcal{T}}^l$ by Algorithm \ref{alg1}. One way is to directly remove these filters following the hard pruning methods \cite{Li2016PruningFF,Molchanov2017Prune} by setting the corresponding filter weight to zeros.
However, the operation may impact the capacity of the well-trained model, especially when a large proportion of filters is dropped. Although several soft pruning methods, \eg, \cite{he2018SFP}, propose to leverage the dynamic mask, which do not drop the filter until the training completion, the side effect caused by setting the filter to zeros also compromises the model training process.
%a large deviation of parameter initialization from the pretrained model, \ie, most of the parameters are set to zeros.
To solve this problem, we propose a simple but effective strategy to gradually reduce the value of candidate filters. Formally, for the $l$-th layer, we set the selected filters $w_i  \leftarrow  \gamma w_i$, \text{where} $ w_i \in \hat{\mathcal{T}}^l , 1\leq i \leq |\hat{\mathcal{T}}^l |, 0 \leq \gamma \leq 1$. The scale of the candidate filter is decreased. Then we fine-tune the network for one epoch. The local geometry score between filters is calculated again, and we re-selected the pruned filters. In other words, the wrong-selected filters are provided more chances to recover the original scale. 
This `decreasing-fine-tune' process is iteratively performed, till the weights of redundant filters gradually converge to zeros. 
%Besides, the parameters of bias and batch normalization of the candidate pruning filters are also decreased. 

\noindent \textbf{Advantages of Filter Weight Decreasing.} 
According to the convolution operation, the filter weight decreasing actually decreases the contribution of the selected filter. For instance, the original output is $w_i \odot x$, and the output of the weight decreasing filter is $ \gamma w_i \odot x$, where $\odot$ denotes the convolution operation. The contribution of the selected filter is also decreased with $\gamma$ times. If $\gamma=0$, the filter weight decreasing will equal to the conventional hard pruning methods. If $\gamma=1$, the network will not be pruned.  The proposed pruning, therefore, could be viewed as one mild strategy of the hard-pruning method. We progressively decrease the contribution of the selected filters, while providing chances of recovering the wrong-selected filters. In Section \ref{sec:parameter}, we further provide the ablation study on the effectiveness of the filter weight decreasing.

\section{Experiments}
\label{sec:experiments}
We apply our pruning method on two sorts of CNN-based retrieval applications, \ie, image retrieval and personal re-identification. The following experiments are conducted with two kinds of networks, \ie, VGG-16 \cite{Simonyan15} and ResNet-50 \cite{heResnet2016}, on three benchmarks, \ie, Oxford5K~\cite{oxford5k2007}, Paris6K~\cite{Paris6k2008}, and Market-1501~\cite{market2015}. 

\begin{figure*}[t]
\begin{minipage}[c]{1\linewidth}
\center{\includegraphics[width=0.9\textwidth]{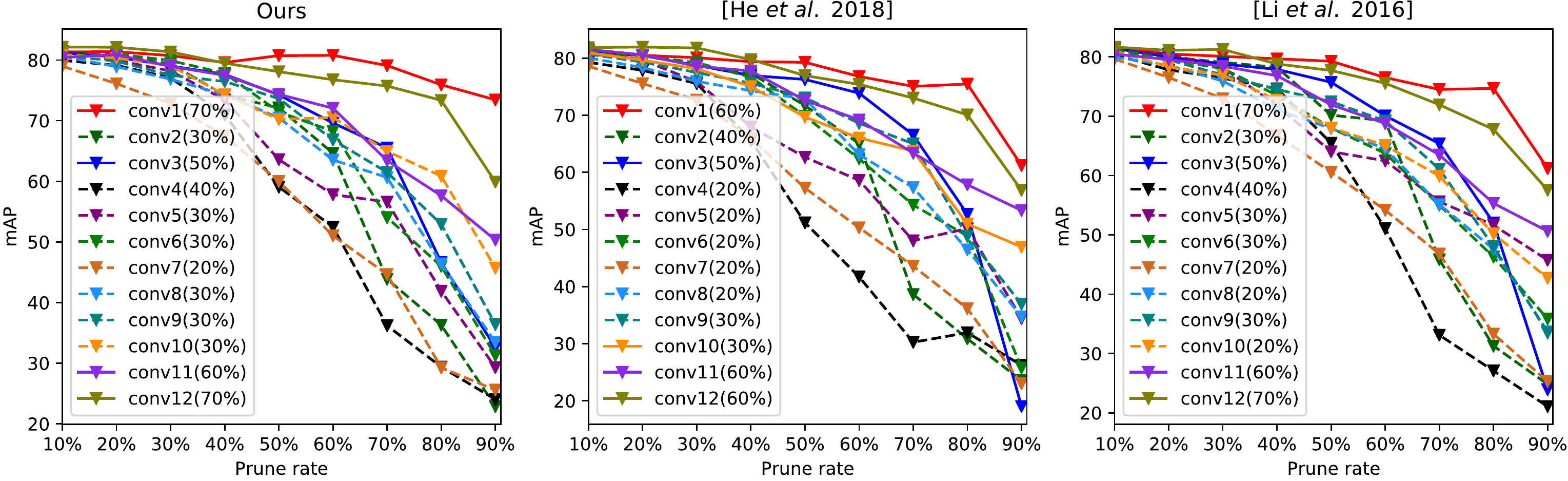}}
\end{minipage} 
\vspace{-.1in}
\caption[]{The pruning sensitivity of each convolutional layer for different pruning methods. The sensitive layers are marked as dashed lines and optimal prune rates for each layer are shown in figure legends. (\textbf{left}: Ours, \textbf{middle}: \cite{he2018SFP}, \textbf{right}: \cite{Li2016PruningFF}). 
}
\label{fig_layer_sensi}
\end{figure*}

\subsection{Results on Personal Re-id}
\label{sec:results_Reid}
\noindent \textbf{Experimental Setting:} For the re-id application, following  \cite{zheng2019joint}, we deploy ResNet-50 pretrained on ImageNet \cite{Imagenet2009} as the backbone network and replace the last average pooling layer and fully-connected layer with an adaptive max-pooling layer. We employ triplet loss function \cite{Tripletloss2016} and SGD to train the network with an initial learning rate $lr=0.001$, momentum 0.9 and the batch size of $32$. We stop training until the $100$-th epoch.

\noindent \textbf{Pruning setting:} Three state-of-the-art methods, \ie,~\cite{he2018SFP}, \cite{he2019FPGM},  ~and~\cite{Li2016PruningFF}  are introduced for comparison. Following \cite{he2018SFP}, all the convolutional layers are pruned with the same pruning rate $P$ at the same time. We test all the comparison pruning methods with $P$ from $\{10\%, 20\%, 50\%, 90\%\}$.  For soft pruning methods, \ie, \cite{he2018SFP}, \cite{he2019FPGM}, and our method, we prune filters while fine-tuning for 100 epochs.  For the hard pruning methods, \ie, \cite{Li2016PruningFF} , to conduct a fair comparison, we prune network once and fine-tune it with the same epochs as soft pruning methods. Our method contains two hyper-parameters, \ie, $k$ and $\gamma$. We show the results with different neighbor numbers $k$ of $\{1,5,10\}$ and set $\gamma = 0.01$ when $P\leqslant 50\%$ and $\gamma = 0.3$ when $P > 50\%$, respectively.   

\noindent \textbf{Results:} Table \ref{table:res50_market} shows the quantitative results. We observe that our method achieves the best performance for almost all the pruning rate $P$. For example, when pruning $50\%$ filters, our method just increases the $top\text{-}5$ error 0.45\%. As the pruning rate increasing, our method obtains much better results than the comparison methods. For instance, our method surpasses other methods over $8\%$ on $mAP$ and $6\%$ on $top\text{-}1$ accuracy respectively, when pruning $90\%$ filters and $88.9\%$ FLOPs\footnote{https://github.com/sovrasov/flops-counter.pytorch} have been reduced. 

\begin{table*}[ht]
\centering
\setlength{\abovecaptionskip}{0pt}
\setlength{\tabcolsep}{4pt}
\small
\caption{Performance evaluation of the compared pruning methods on  Market-1501 using ResNet-50 initialized on the Imagenet dataset. }
\label{table:res50_market}

\newcommand{\tabincell}[2]{\begin{tabular}{@{}#1@{}}#2\end{tabular}}
\begin{tabular}{c| c c c c c c c }
\shline
 Pruning rate(\%)   & Methods &  mAP(\%) & Rank@1(\%) &Rank@5(\%) &Rank@10(\%) &FLOPs(\%)$\downarrow$& Parameters(\%)$\downarrow$    \\
 \shline 
 0 & \cite{zheng2019joint}&    70.81  &86.63 &93.74 &96.05  &0 & 0 \\\hline 
 
\multirow{6}*{10}  & \cite{he2018SFP} &69.08 &85.45 &93.85 &95.78& \multirow{6}*{14.50} &\multirow{6}*{12.92} \\
   & \cite{he2019FPGM}& 70.23 &86.40 &93.97 &95.81   & & \\
   & \cite{Li2016PruningFF} &69.08 & 85.33 &93.88 &96.17 &  & \\
   & Ours (k=1)& 70.06 & 86.22 &94.18 &95.99 &  &   \\
   & Ours (k=5)& 70.38 & 86.25 &94.18 &96.26 &  &   \\
   & Ours (k=10)& \textbf{70.53} & \textbf{86.58} &\textbf{94.39} &\textbf{96.29} &  &   \\\hline

\multirow{6}*{20} & \cite{he2018SFP} &   69.03 &84.77 &93.74 &95.69 &\multirow{6}*{28.25} &\multirow{6}*{24.83}  \\
  & \cite{he2019FPGM} & 70.07 &\textbf{86.07} &93.79 &95.81 &  &  \\
  &\cite{Li2016PruningFF}  & 67.93 &85.51 &\textbf{93.97} &\textbf{96.14} &  & \\
  & Ours (k=1)& 69.60 &85.60 &93.91 &96.08 &  &  \\
  & Ours (k=5)& 69.76 &85.78 &93.94 &95.93 &  &   \\
  & Ours (k=10)& \textbf{70.23} &\textbf{86.07} &93.59 &95.69 &  &  \\\hline

 \multirow{6}*{50} & \cite{he2018SFP}&  65.85 & 83.05 &92.96 &95.52 &\multirow{6}*{62.45} &\multirow{6}*{55.57}   \\
  & \cite{he2019FPGM} & 65.31 &83.02 &91.98 &94.66 &  &    \\
  & \cite{Li2016PruningFF} &57.22 &77.82 &90.29 &93.74 &  & \\
  &  Ours (k=1)& \textbf{66.32} &83.85 &\textbf{93.29} &\textbf{95.64} &  &    \\
  &  Ours (k=5)& 66.14 &83.64 &92.84 &94.93 &  &    \\
  &  Ours (k=10)& 66.09 &\textbf{83.88} &93.11 &95.55 &  &    \\\hline

 \multirow{6}*{90} & \cite{he2018SFP}& 48.02 &71.17 &86.88 &90.91 &\multirow{6}*{88.85} &\multirow{6}*{74.28}  \\
  & \cite{he2019FPGM}& 45.31 &68.26 &84.74 &89.64 &  &  \\
  & \cite{Li2016PruningFF} & 47.24 &70.57 &85.09 &89.85 &  &   \\
  & Ours (k=1)& \textbf{56.47} &\textbf{77.25} &\textbf{89.46} &\textbf{92.84} &  &   \\
  & Ours (k=5)&50.98 &72.06 &87.11 &91.33 &  &   \\
  & Ours (k=10)&49.29 &70.31 &86.40 &91.24 &  &   \\
\shline
\end{tabular}
\end{table*}

\subsection{Results on Image Retrieval}

\noindent \textbf{Setting:} We deploy the same backbone network, \ie, VGG-16, as \cite{Radenovi2018Retrieval} for image retrieval. %which is based on VGG-16. %, and skip the whitening operation for simplicity. 
We evaluate the pruning methods on models pretrained on the building dataset, \ie, Structure-from-Motion 3D (SfM3D)~\cite{SfmDataset2015}. The pretrained model on SfM3D is provided by the author \footnote{https://github.com/filipradenovic/cnnimageretrieval-pytorch}. We follow the conventional pipeline in \cite{Radenovi2018Retrieval} and extract the multi-scale representations from the images of different scale factors, \ie, $\{\frac{1}{\sqrt{2}},1,\sqrt{2} \}$. We adopt the contrastive loss function as the fine-tuning objective and use Adam \cite{kingma2014adam} to fine-tune the model with an initial learning rate $l_0=5\times 10^{-6}$, exponential decay $lr = l_0\exp{(-0.01i)}$  over epoch $i$, momentum $0.9$, margin $0.85$, and the batch size of $5$. 

\noindent \textbf{Pruning setting:} Three state-of-the-art methods are chosen for comparison, \ie, \cite{Li2016PruningFF}, \cite{Molchanov2017Prune}, and \cite{he2018SFP}. Following \cite{Li2016PruningFF}, we investigate the pruning sensitivity of each convolutional layer and manually choose the best pruning rates (see Fig.~\ref{fig_layer_sensi}). For \cite{Molchanov2017Prune}, which ranks filters through all convolution layers, we set the pruning rate to $0.4$. For our method, the hyper-parameters $k$ and $\gamma$ are set to 10 and 0.6, respectively.

\noindent \textbf{Results:} Table \ref{table:vgg_ox5k} summarizes the comparison results on the Oxford5K and Paris6K datasets. The proposed method arrives competitive results $76.20\%$ mAP on Oxford5K, while the $57.37\%$ FLOPs are reduced. The performance drop is limited with $-6.25\%$ mAP. Similar results are observed on Paris6K. When $61.05\%$ parameters are reduced, the pruned model still arrives a competitive performance with other pruning methods, which verifies the effectiveness of our method.

\begin{table}[ht]
\centering
\setlength{\abovecaptionskip}{0pt}
\setlength{\tabcolsep}{7pt}
\caption{Performance evaluation of the compared pruning methods using VGG-16 on  Oxford5K and  Paris6K. %`I' and `S' represent networks initialized on ImageNet or the SfM3D dataset, respectively
}
\label{table:vgg_ox5k}
\setlength\dashlinedash{0.2pt}
\setlength\dashlinegap{1.5pt}
\setlength\arrayrulewidth{0.3pt}
\newcommand{\tabincell}[2]{
\begin{tabular}{@{}#1@{}}#2\end{tabular}}
\resizebox{\linewidth}{!}{
\begin{tabular}{l| c  c c c   }
\shline 
 Dataset & Methods & mAP(\%) & FLOPs(\%)$\downarrow$& Parameters(\%)$\downarrow$   \\\shline 
 
\multirow{5}*{Oxford5K} &  \cite{Radenovi2018Retrieval} &    82.45  &- & -  \\
  &\cite{Li2016PruningFF}  & 75.25 & 50.70 & 55.61  \\ %\cdashline{2-6}
& \cite{Molchanov2017Prune} & 76.15   & 52.39  & 70.43   \\ %\cdashline{2-6}
& \cite{he2018SFP} & 75.26&  51.59 & 57.38 \\ %\cdashline{2-6}
& Ours  &\textbf{76.20} & 57.37 & 61.05 \\\hline

\multirow{5}*{Paris6K} &  \cite{Radenovi2018Retrieval} &    81.37  &- & -  \\
 &\cite{Li2016PruningFF}    & 71.78  & 50.70  & 55.61  \\%\cdashline{2-6}
& \cite{Molchanov2017Prune} & 72.51 & 52.39 & 70.43  \\%\cdashline{2-6}
& \cite{he2018SFP} & \textbf{74.02} & 51.59  & 57.38 \\%\cdashline{2-6}
& Ours & 73.18  & 57.37 & 61.05  \\
\shline
\end{tabular}
}
\end{table}

\begin{figure}[t]
\centering{
\begin{minipage}[c]{0.8\linewidth}
\includegraphics[width=\textwidth]{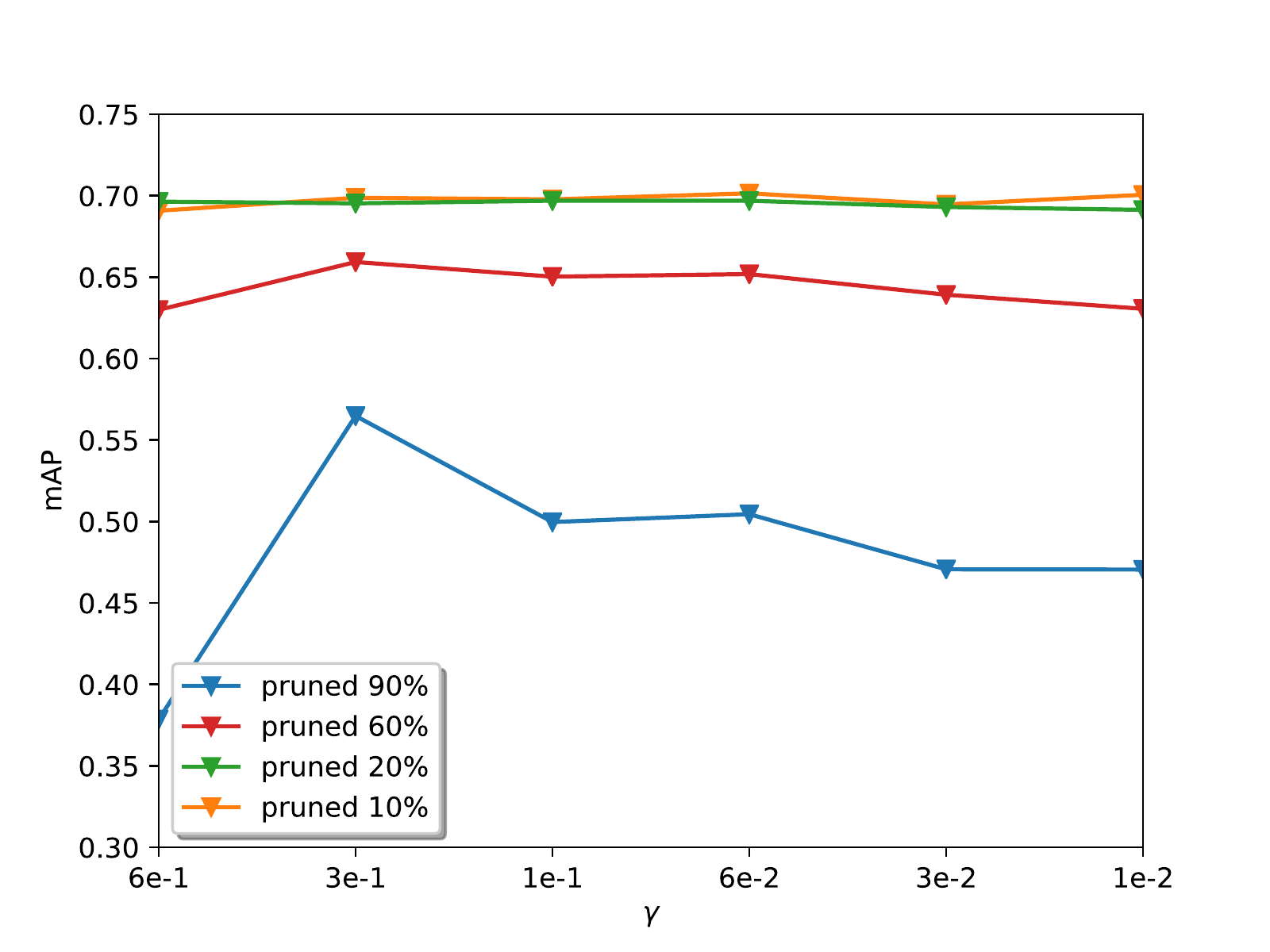}
\end{minipage} }
%\vspace{-.1in}
\caption{The parameter sensitivity analysis of our method on the Market-1501 dataset.}
\label{fig_para_sensi}
\end{figure}

\subsection{Parameter Sensitivity} \label{sec:parameter}

\noindent The proposed method contains two hyper-parameters, \ie, the number of neighbors $k$ and the weight decreasing level $\gamma$. %A larger $k$ indicates a wider spread of neighbors for local power calculation. Increasing $\gamma$ will slow down the  weight changing for redundant filters.
A large $k$ indicates that more neighbor filters are taken into consideration. Meanwhile, the value of $\gamma$ affects the speed of the weight decreasing.

\noindent \textbf{How to determine the value of $k$ ?} From Table.\ref{table:res50_market}, we could observe that our method is sensitive to the parameter $k$. Concretely, if $P$ is relatively small, \eg, $P \leq 20\%$, a large $k$ value will lead to a better performance. In contrast, a small $k$ value is the first choice for the high-proportion pruning demand, \eg, $P \geq 50\%$. We speculate that when a small proportion of filters is selected, a large $k$ could utilize more local information without damaging the original filter structure. When we intend to remove a large proportion of filters, the large $k$ may lead to dropping the entire points in the local geometry, which could be avoided by searching the limited local geometry. A small $k$ value, therefore, performs well in this condition. Only the closest neighbor is taken into consideration.

\noindent \textbf{The effect of weight decreasing.} To study the importance of parameter $\gamma$, we evaluate different  $\gamma$ values from $\{0.6, 0.3, 0.1, 0.06, 0.03, 0.01\}$ on the Market-1501 dataset under different pruning rates $P$ varying from $\{10\%, 20\%, 60\%, 90\%\}$. We fix $k=1$ in the ablation study. The results are shown in Fig.~\ref{fig_para_sensi}. We observe that our method is not sensitive to $\gamma$ when $P$ is small, \eg, $P =10\% ~\text{and}~ P=20\%$. In contrast, when a large proportion of parameters is dropped, \eg, $P > 50\%$, the larger $\gamma$ significantly performs well, which decreases the filter scale slowly. It is consistent with our intuition that the filter weight decreasing helps the model to adapt to the large prune rate. 
Note that if the value of $\gamma$ is too large, \eg, $\gamma=0.6$, the network may converge very slowly, and limited filters are decreased to zeros. To compare the results fairly, we use the hard-pruning method to drop the final weights, so the model of $\gamma = 0.6$ does not perform well.

\subsection{Feature Map Visualization} 
\begin{figure}[t]
\center{
\subfigure{
\begin{minipage}[c]{0.18\linewidth}
\center{\includegraphics[width=1\textwidth,height=0.145\textheight]{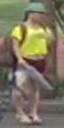}}
\end{minipage} 
}
\subfigure{
\begin{minipage}[c]{0.75\linewidth}
\center{\includegraphics[width=\textwidth]{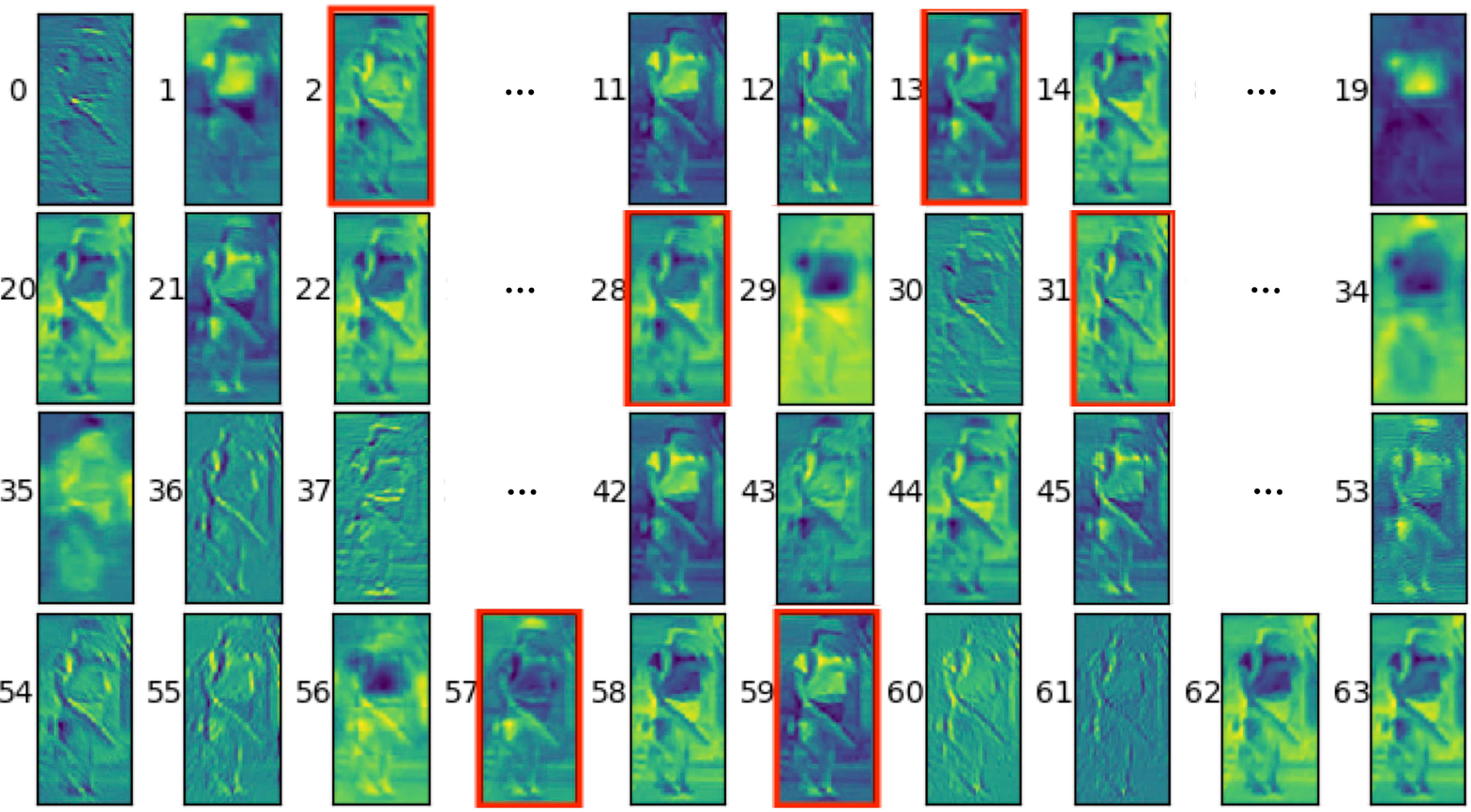}}
\end{minipage} 
}
}
\vspace{-.1in}
\caption{The input image (\textbf{left}) and the visualization of the first convolutional layer feature maps (indexed from 0 to 63) (\textbf{right}). The maps with red boxes are pruned by our method.% with a 10\% pruning rate.
}
\vspace{-.1in}
\label{fig_fea_map}
\end{figure}

To explore the effectiveness of the proposed method, 
we further visualize the first convolutional layer feature maps shown in Fig.~\ref{fig_fea_map}, where we have pruned 10\% of filters and marked the corresponding feature maps with red boxes. These pruned feature maps contain the outlines of the input image, such as straps~(2, 31), T-shirts~(13, 59), shorts~(28), and hat~(57). Clearly, these feature maps can be replaced by the remaining ones. For example, straps, T-shirts, shorts, and hats can be replaced by feature maps: (43, 54, 55,~\etal{}), (11, 42, ~\etal{}), (20, 22 ~\etal{}), and (1, 34, 53, ~\etal{}) respectively.  

\subsection{Embedding Feature Analysis}
\begin{table}[t]
\centering
\small
\setlength{\abovecaptionskip}{0pt}
\caption{The mean Euclidean distance on Market-1501 between the embedding features extracted from the original model and the pruned models with different pruning methods. }
\label{table:fea_distance}
\begin{tabular}{c| c |c}
\shline
 Methods & Distance~( $\times 10^{-2}$) & mAP(\%) \\
 \shline
 \cite{he2019FPGM}& 35.28 &45.31 \\
 \cite{Li2016PruningFF} &35.17 &47.24 \\
 \cite{he2018SFP} & 35.13 &48.02 \\
 Ours ($k=1$) &34.70 &56.47  \\
\shline
\end{tabular}
\end{table}

As mentioned above, maintaining the original filter distribution  of the network is critical for image retrieval. To study it, we compute the Euclidean distance between the embedding features of the original model and pruned models with 90\% filters removed. Our model extracts the most similar features with the baseline model, which indicates that the proposed method is suitable for pruning image retrieval networks. 

\section{Conclusion}
 This paper has proposed a progressive local filter pruning strategy for image retrieval acceleration. Different from the existing global center-based pruning methods, our method seeks to reduce filter redundancy by considering the local relations of filters. Furthermore, the filter weight decreasing allows the pruned model to recover the representation capability, especially when dropping a large proportion of filters. Compelling results on computation reduction and feature representation maintaining demonstrate the effectiveness of our method.

{\footnotesize
\bibliographystyle{named}
\bibliography{ijcai20}

\begin{thebibliography}{}

\bibitem[\protect\citeauthoryear{{Arandjelović} \bgroup \em et al.\egroup
  }{2018}]{Tripletloss2016}
R.~{Arandjelović}, P.~{Gronat}, A.~{Torii}, T.~{Pajdla}, and J.~{Sivic}.
\newblock Netvlad: Cnn architecture for weakly supervised place recognition.
\newblock {\em TPAMI}, 2018.

\bibitem[\protect\citeauthoryear{Babenko and
  Lempitsky}{2015}]{Babenko2015AggregatingLD}
Artem Babenko and Victor~S. Lempitsky.
\newblock Aggregating local deep features for image retrieval.
\newblock {\em ICCV}, 2015.

\bibitem[\protect\citeauthoryear{Babenko \bgroup \em et al.\egroup
  }{2014}]{Babenko2014NeuralCF}
Artem Babenko, Anton Slesarev, Alexander Chigorin, and Victor~S. Lempitsky.
\newblock Neural codes for image retrieval.
\newblock abs/1404.1777, 2014.

\bibitem[\protect\citeauthoryear{Chen \bgroup \em et al.\egroup
  }{2016}]{Chen2016KnowledgeTransfer}
Tianqi Chen, Ian Goodfellow, and Jonathon Shlens.
\newblock Net2net: Accelerating learning via knowledge transfer.
\newblock In {\em ICLR}, 2016.

\bibitem[\protect\citeauthoryear{{Deng} \bgroup \em et al.\egroup
  }{2009}]{Imagenet2009}
J.~{Deng}, W.~{Dong}, R.~{Socher}, L.~{Li}, {Kai Li}, and {Li Fei-Fei}.
\newblock Imagenet: A large-scale hierarchical image database.
\newblock In {\em CVPR}, 2009.

\bibitem[\protect\citeauthoryear{Frankle and
  Carbin}{2018}]{Jonathan2018Lottery}
Jonathan Frankle and Michael Carbin.
\newblock The lottery ticket hypothesis: Training pruned neural networks.
\newblock {\em CoRR}, abs/1803.03635, 2018.

\bibitem[\protect\citeauthoryear{Gong \bgroup \em et al.\egroup
  }{2014}]{Gong2014CompressingDC}
Yunchao Gong, Liu Liu, Ming Yang, and Lubomir~D. Bourdev.
\newblock Compressing deep convolutional networks using vector quantization.
\newblock abs/1412.6115, 2014.

\bibitem[\protect\citeauthoryear{Han \bgroup \em et al.\egroup
  }{2015a}]{Han2015DeepCC}
Song Han, Huizi Mao, and William~J. Dally.
\newblock Deep compression: Compressing deep neural network with pruning,
  trained quantization and huffman coding.
\newblock abs/1510.00149, 2015.

\bibitem[\protect\citeauthoryear{Han \bgroup \em et al.\egroup
  }{2015b}]{HanPTD2015}
Song Han, Jeff Pool, John Tran, and William~J. Dally.
\newblock Learning both weights and connections for efficient neural networks.
\newblock abs/1506.02626, 2015.

\bibitem[\protect\citeauthoryear{Hassibi \bgroup \em et al.\egroup
  }{1994}]{NIPHassibiS1993}
Babak Hassibi, David~G. Stork, and Gregory Wolff.
\newblock Optimal brain surgeon: Extensions and performance comparisons.
\newblock In {\em NeurIPS}. 1994.

\bibitem[\protect\citeauthoryear{{He} \bgroup \em et al.\egroup
  }{2016}]{heResnet2016}
K.~{He}, X.~{Zhang}, S.~{Ren}, and J.~{Sun}.
\newblock Deep residual learning for image recognition.
\newblock In {\em CVPR}, 2016.

\bibitem[\protect\citeauthoryear{He \bgroup \em et al.\egroup
  }{2017}]{He2017ChannelPF}
Yihui He, Xiangyu Zhang, and Jian Sun.
\newblock Channel pruning for accelerating very deep neural networks.
\newblock {\em ICCV}, 2017.

\bibitem[\protect\citeauthoryear{He \bgroup \em et al.\egroup
  }{2018}]{he2018SFP}
Yang He, Guoliang Kang, Xuanyi Dong, Yanwei Fu, and Yi~Yang.
\newblock Soft filter pruning for accelerating deep convolutional neural
  networks.
\newblock In {\em IJCAI}, 2018.

\bibitem[\protect\citeauthoryear{He \bgroup \em et al.\egroup
  }{2019}]{he2019FPGM}
Yang He, Ping Liu, Ziwei Wang, Zhilan Hu, and Yi~Yang.
\newblock Filter pruning via geometric median for deep convolutional neural
  networks acceleration.
\newblock In {\em CVPR}, 2019.

\bibitem[\protect\citeauthoryear{Hinton \bgroup \em et al.\egroup
  }{2015}]{DistrillNIPS2015}
Geoffrey Hinton, Oriol Vinyals, and Jeffrey Dean.
\newblock Distilling the knowledge in a neural network.
\newblock In {\em NIPS Deep Learning and Representation Learning Workshop},
  2015.

\bibitem[\protect\citeauthoryear{Jaderberg \bgroup \em et al.\egroup
  }{2014}]{Jaderberg2014}
Max Jaderberg, Andrea Vedaldi, and Andrew Zisserman.
\newblock Speeding up convolutional neural networks with low rank expansions.
\newblock In {\em BMVC}, 2014.

\bibitem[\protect\citeauthoryear{Kim \bgroup \em et al.\egroup
  }{2018}]{DistrillNIPS2018}
Jangho Kim, Seonguk Park, and Nojun Kwak.
\newblock Paraphrasing complex network: Network compression via factor
  transfer.
\newblock In {\em NeurIPS}. 2018.

\bibitem[\protect\citeauthoryear{Kingma and Ba}{2014}]{kingma2014adam}
Diederik~P Kingma and Jimmy Ba.
\newblock Adam: A method for stochastic optimization.
\newblock {\em arXiv:1412.6980}, 2014.

\bibitem[\protect\citeauthoryear{LeCun \bgroup \em et al.\egroup
  }{1990}]{LeCunNIPS1989}
Yann LeCun, John~S. Denker, and Sara~A. Solla.
\newblock Optimal brain damage.
\newblock In {\em NeurIPS}. 1990.

\bibitem[\protect\citeauthoryear{Li \bgroup \em et al.\egroup
  }{2016}]{Li2016PruningFF}
Hao Li, Asim Kadav, Igor Durdanovic, Hanan Samet, and Hans~Peter Graf.
\newblock Pruning filters for efficient convnets.
\newblock abs/1608.08710, 2016.

\bibitem[\protect\citeauthoryear{Li \bgroup \em et al.\egroup
  }{2017}]{Li2017ijcai}
Xuelong Li, Mulin Chen, Feiping Nie, and Qi~Wang.
\newblock Locality adaptive discriminant analysis.
\newblock In {\em IJCAI}, 2017.

\bibitem[\protect\citeauthoryear{Pavlo \bgroup \em et al.\egroup
  }{2017}]{Molchanov2017Prune}
Molchanov Pavlo, Tyree Stephen, Karras Tero, Aila Timo, and Kautz Jan.
\newblock Pruning convolutional neural networks for resource efficient
  inference.
\newblock {\em ICLR}, 2017.

\bibitem[\protect\citeauthoryear{{Philbin} \bgroup \em et al.\egroup
  }{2007}]{oxford5k2007}
J.~{Philbin}, O.~{Chum}, M.~{Isard}, J.~{Sivic}, and A.~{Zisserman}.
\newblock Object retrieval with large vocabularies and fast spatial matching.
\newblock In {\em CVPR}, 2007.

\bibitem[\protect\citeauthoryear{{Philbin} \bgroup \em et al.\egroup
  }{2008}]{Paris6k2008}
J.~{Philbin}, O.~{Chum}, M.~{Isard}, J.~{Sivic}, and A.~{Zisserman}.
\newblock Lost in quantization: Improving particular object retrieval in large
  scale image databases.
\newblock In {\em CVPR}, 2008.

\bibitem[\protect\citeauthoryear{Radenovi{\'c} \bgroup \em et al.\egroup
  }{2018}]{Radenovi2018Retrieval}
F.~Radenovi{\'c}, G.~Tolias, and O.~Chum.
\newblock Fine-tuning {CNN} image retrieval with no human annotation.
\newblock {\em TPAMI}, 2018.

\bibitem[\protect\citeauthoryear{Rastegari \bgroup \em et al.\egroup
  }{2016}]{rastegariECCV16}
Mohammad Rastegari, Vicente Ordonez, Joseph Redmon, and Ali Farhadi.
\newblock Xnor-net: Imagenet classification using binary convolutional neural
  networks.
\newblock In {\em ECCV}, 2016.

\bibitem[\protect\citeauthoryear{Sch{\"o}nberger \bgroup \em et al.\egroup
  }{2015}]{SfmDataset2015}
Johannes~L. Sch{\"o}nberger, Filip Radenovi{\'c}, Ondřej Chum, and Jan-Michael
  Frahm.
\newblock From single image query to detailed 3d reconstruction.
\newblock {\em CVPR}, 2015.

\bibitem[\protect\citeauthoryear{Simonyan and Zisserman}{2015}]{Simonyan15}
Karen Simonyan and Andrew Zisserman.
\newblock Very deep convolutional networks for large-scale image recognition.
\newblock In {\em ICLR}, 2015.

\bibitem[\protect\citeauthoryear{Sun \bgroup \em et al.\egroup
  }{2019}]{sun2019learning}
Tianxiang Sun, Yunfan Shao, Xiaonan Li, Pengfei Liu, Hang Yan, Xipeng Qiu, and
  Xuanjing Huang.
\newblock Learning sparse sharing architectures for multiple tasks.
\newblock {\em AAAI}, 2019.

\bibitem[\protect\citeauthoryear{Tolias \bgroup \em et al.\egroup
  }{2016}]{toliasICLR2016}
Giorgos Tolias, Ronan Sicre, and Herv{\'e} J{\'e}gou.
\newblock {Particular Object Retrieval With Integral Max-Pooling of CNN
  Activations}.
\newblock In {\em ICLR}, 2016.

\bibitem[\protect\citeauthoryear{Yang \bgroup \em et al.\egroup
  }{2011}]{Yang2011UDFS}
Yi~Yang, Heng~Tao Shen, Zhigang Ma, Zi~Huang, and Xiaofang Zhou.
\newblock L2,1-norm regularized discriminative feature selection for
  unsupervised learning.
\newblock In {\em IJCAI}, 2011.

\bibitem[\protect\citeauthoryear{{Zheng} \bgroup \em et al.\egroup
  }{2015}]{market2015}
L.~{Zheng}, L.~{Shen}, L.~{Tian}, S.~{Wang}, J.~{Wang}, and Q.~{Tian}.
\newblock Scalable person re-identification: A benchmark.
\newblock In {\em ICCV}, 2015.

\bibitem[\protect\citeauthoryear{Zheng \bgroup \em et al.\egroup
  }{2016}]{ZhengAxiv16}
Liang Zheng, Yi~Yang, and Alexander~G. Hauptmann.
\newblock Person re-identification: Past, present and future.
\newblock {\em CoRR}, abs/1610.02984, 2016.

\bibitem[\protect\citeauthoryear{Zheng \bgroup \em et al.\egroup
  }{2017}]{ZhengTOMM2017}
Zhedong Zheng, Liang Zheng, and Yi~Yang.
\newblock A discriminatively learned cnn embedding for person reidentification.
\newblock {\em ACM TOMM}, 2017.

\bibitem[\protect\citeauthoryear{Zheng \bgroup \em et al.\egroup
  }{2019}]{zheng2019joint}
Zhedong Zheng, Xiaodong Yang, Zhiding Yu, Liang Zheng, Yi~Yang, and Jan Kautz.
\newblock Joint discriminative and generative learning for person
  re-identification.
\newblock {\em CVPR}, 2019.

\end{thebibliography}
}

\end{document}